%% file: root.tex
\documentclass[letterpaper, 10 pt, conference, table]{ieeeconf}  %

\IEEEoverridecommandlockouts                              %

\usepackage{graphics} %
\usepackage{algorithm2e}
\usepackage{todonotes}

\usepackage{amsmath}
\usepackage{amssymb}
\usepackage{booktabs}
\usepackage[]{graphicx}
\usepackage[font=footnotesize]{caption}
\usepackage[font=footnotesize]{subcaption}
\usepackage{mathtools}
\usepackage{gensymb}
\usepackage{lipsum}
\usepackage{float}
\usepackage{color}
\usepackage{caption}
\usepackage{threeparttable}
\usepackage{todonotes}
\usepackage{mathtools}
\usepackage[]{algorithm2e}
\usepackage{soul}
\usepackage{siunitx}
\usepackage{tabularx}

\newcommand\MYhyperrefoptions{bookmarks=true,bookmarksnumbered=true,
pdfpagemode={UseOutlines},plainpages=false,pdfpagelabels=true,
colorlinks=true,citecolor={black},
pdftitle={Voxel Map for Visual SLAM},%
pdfsubject={Computer Vision, Robotics},%
pdfauthor={M. Muglikar, Z. Zhang, D. Scaramuzza},%
pdfkeywords={SLAM, Multi-Camera}}%

\DeclarePairedDelimiter{\round}\lfloor\rceil

\newcommand{\Sec}{Section~}
\newcommand{\Fig}{Fig.~}

\newcommand{\red}[1]{\textcolor{red}{#1}}
\newcommand{\eg}{e.g., }
\newcommand{\ie}{i.e., }
\newcommand{\etal}{et al. }

\newcommand{\improv}[1]{{\color{blue}#1}}

\usepackage[\MYhyperrefoptions, pdftex, pagebackref=true,breaklinks=true,letterpaper=true,colorlinks,bookmarks=false]{hyperref} 
\definecolor{somegray}{rgb}{0.5, 0.5, 0.5}
\newcommand{\darkgrayed}[1]{\textcolor{somegray}{#1}}
\makeatletter
\newcommand*\titleheader[1]{\gdef\@titleheader{#1}}
\AtBeginDocument{%
  \let\st@red@title\@title
  \def\@title{%
    \vskip-3em
    \bgroup\normalfont\large\centering\@titleheader\par\egroup
    \vskip1.5em\st@red@title}
}
\makeatother

\titleheader{\darkgrayed{This paper has been accepted for publication at the IEEE International Conference on Robotics and Automation (ICRA), Paris, 2020.
\copyright IEEE}}

\title{\LARGE \bf
Voxel Map for Visual SLAM
}

\author{Manasi Muglikar, Zichao Zhang and Davide Scaramuzza%
\thanks{The authors are with the Robotics and Perception Group, Dep. of Informatics, University of Zurich , and Dep. of Neuroinformatics, University of Zurich and ETH Zurich, Switzerland--- \url{http://rpg.ifi.uzh.ch.}
This research was supported by the National Centre of Competence in Research (NCCR) Robotics, through the Swiss National Science Foundation, the SNSF-ERC Starting Grant and Sony R\&D Center Europe.}%
}

\begin{document}

\maketitle
\thispagestyle{empty}
\pagestyle{empty}

\begin{abstract}
In modern visual SLAM systems, it is a standard practice to retrieve potential candidate map points from overlapping keyframes for further feature matching or direct tracking.
In this work, we argue that keyframes are not the optimal choice for this task, due to several inherent limitations, such as weak geometric reasoning and poor scalability.
We propose a voxel-map representation to efficiently retrieve map points for visual SLAM.
In particular, we organize the map points in a regular voxel grid.
Visible points from a camera pose are queried by sampling the camera frustum in a raycasting manner, which can be done in constant time using an efficient voxel hashing method.
Compared with keyframes, the retrieved points using our method are geometrically guaranteed to fall in the camera field-of-view, and occluded points can be identified and removed to a certain extend.
This method also naturally scales up to large scenes and complicated multi-camera configurations.
Experimental results show that our voxel map representation is as efficient as a keyframe map with $5$ keyframes and provides significantly higher localization accuracy (average 46\% improvement in RMSE) on the EuRoC dataset.
The proposed voxel-map representation is a general approach to a fundamental functionality in visual SLAM and widely applicable.

\end{abstract}

\input{sections/introduction.tex}
\input{sections/voxelhashing.tex}

\input{sections/systemoverview.tex}

\input{sections/experiments.tex}

\input{sections/conclusion.tex}

\IEEEtriggeratref{12}
\bibliographystyle{IEEEtran}
\bibliography{all_rpg,references}

\end{document}

%% file: sections/introduction.tex
\section{introduction}\label{intro}
Simultaneous Localization and Mapping (SLAM) is fundamental to robotics and plays a pivotal role in various real-world applications, such as augmented/virtual reality and autonomous driving.
The past decade has witnessed rapid progress in this field.
Today, state-of-the-art SLAM systems, specifically visual-inertial SLAM, execute in real-time on power and memory constrained devices and provide accurate and robust estimate.
Despite the remaining challenges in this field \cite{Cadena16tro}, SLAM has reached the maturity that enables successful commercial applications (\eg \cite{oculus_quest}).
Keyframe-based SLAM, among other paradigms such as filter-based method, is arguably the most successful one nowadays.
In particular, keyframe-based SLAM relies on the joint nonlinear optimization of keyframes and visible landmarks, namely bundle adjustment (BA) \cite{Triggs00}, and achieves superior accuracy than filter-based methods \cite{Strasdat10icra}.

Following seminal work of \cite{Klein07ismar}, most state-of-art  sparse SLAM systems use parallel threads for tracking (\ie compute real-time poses for image stream) and BA to alleviate the computational overhead of the nonlinear optimization.
The central task of the tracking process, for both direct and feature-based methods, is to find 2D-3D correspondences between the observations in the current image and the map (\eg 3D points).
While different types of maps are used for dense SLAM (\eg Truncated Signed Distance Field in \cite{Newcombe11ismar}, surfels in \cite{Whelan15rss}), little work has been done exploring alternative map representations for sparse SLAM.
Sparse keyframe-based methods use information from nearby-keyframes to associate images to map points.
This is a powerful heuristic that is used in many successful systems \cite{Klein07ismar, MurArtal15tro, Forster17troSVO, Engel17pami}.
Another class of SLAM systems use geometric primitives (\eg meshes~\cite{Rosinol19icra} or planes~\cite{Kaess15icra,Nardi19ral})

\begin{figure}[t!]
    \centering
    \includegraphics[width=\linewidth]{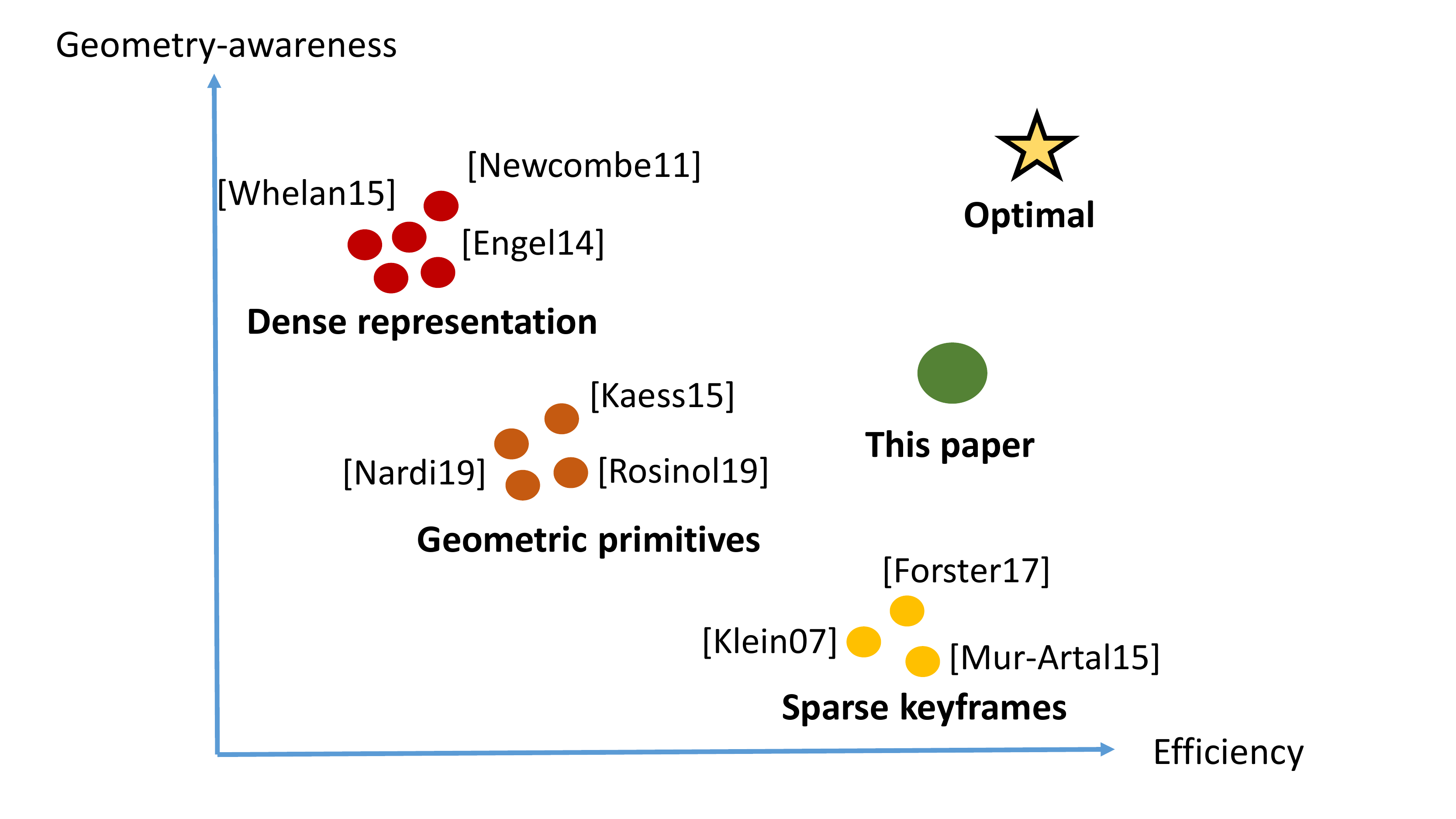}
    \caption{The optimal SLAM systems should be efficient and have geometrical understanding of the map (denoted by the golden star).
    Direct methods (red dots) %
    associate each keyframe with a semi-dense depth map. %
    They have more scene information but are not efficient.
    Sparse keyframe-based SLAM (yellow dots) %
    associate features in the current frame to 3D points from nearby overlapping keyframes. 
    While they are computationally efficient, they do not provide higher level understanding of the geometry of the scene.
    Other representations using geometric primitives (orange dots) %
    balance geometric information and efficiency, but make assumptions on the scene and do not achieve efficiency as sparse-keyframe methods. 
    This paper proposes a voxel-map for sparse SLAM, that tries to move one step towards optimal map representations for SLAM. }
    \label{fig:loc_axes}
\end{figure}

Ideally, the map representation should,
(i) be aware of the geometry of the scene 
and (ii) be efficient in terms of computation time and memory. \Fig\ref{fig:loc_axes} shows how different map representations perform on these axes. 
The ideal representation should allow better geometric reasoning, which brings higher accuracy, but still be at par with keyframe-based methods in terms of efficiency.
We compare the effectiveness of sparse keyframe-based map representations along these axes:

\noindent
\textbf{Geometry-awareness}: In sparse SLAM, using keyframes and their visible points (\ie covisibility graph) as the map only allows limited geometric reasoning. 
The co-visibility graph has no notion of occlusion, and it is difficult to determine and filter occluded points, which may cause wrong data association and erroneous estimation. 
It is ideal that the points retrieved from the map coincide with the field-of-view (FoV) of the camera. Unfortunately, there is little geometric guarantee for the points from overlapping keyframes. There may be false positives and missing points.

\noindent
\textbf{Efficiency}:
The effectiveness of keyframes in bundle adjustment comes from the fact that they retain most of the information compared with using all the frames \cite{Strasdat10icra}.
For a local map of $N$ keyframes, increasing $N$ will improve the robustness in general but will result in a longer query time, despite the fact that we are interested in a spatial area of \textit{fixed size} (\ie the camera frustum).
Moreover, the design of keyframe systems becomes complicated for non-trivial multiple camera systems.
For example, setting the images from all cameras as keyframes retain the most information but introduces high redundancy.

Therefore, we argue that using keyframes is not optimal for data association in the tracking process, despite the temptation of using one common representation for different tasks(\ie BA and point retrieval).
In view of the above problems, an ideal map representation for SLAM should be designed for efficient, accurate, geometry-aware points retrieval, rather than simply reusing the keyframes from BA.

In this work, we propose a voxel-map representation that is \textit{scalable} and \textit{geometry-aware}.
By representing the environment as voxels, the map coverage can be specified directly, instead of implicitly depending on keyframe parameters.
Retrieving points from the map is equivalent to accessing the voxels in the area of interest, and the voxel hashing method proposed in \cite{Niessner13tog} allows \textit{constant query time} for a fixed camera frustum (\ie fixed number of voxels), regardless of the number of voxels/points in the map.
Moreover, since voxels are simply containers for 3D points, modifying the information in the voxel-map (\eg adding points from a newly added keyframe) is trivial. %
To query candidate points for data association in SLAM, we propose a raycasting-based method. In particular, we raycast selected pixels from a regular grid in the image to the map and collect the points in the voxels along the rays.
Despite its simplicity, this method has two key advantages.
First, the points returned by our raycasting method are guaranteed to be all the 3D points in the map that fall in the FoV of the camera, for which keyframe-based methods can only rely on the weak covisibility assumption (see \Fig\ref{fig:mh01_vis}).
Second, once we have encountered sufficient 3D points in the nearby voxels along the ray, we can stop checking farther voxels. This gives our method the ability to reason about occlusion to a certain extent. Arguably, this does not provide a complete geometric reasoning as a dense model but is much more efficient.
To the best of our knowledge, this is the first work aiming at incorporating a voxel-based representation in sparse SLAM systems.
As a general approach, our voxel-map representation can be used as a build block for a wide range of SLAM systems.

The rest of the paper is structured as follows.
\Sec\ref{data_struct} introduces the voxel map representation for SLAM and describes the raycasting-based query method using the proposed voxel-hashing based map.
\Sec\ref{algo} demonstrates how the proposed method can be integrated into a modern SLAM pipeline.
The evaluation of the proposed map representation and the comparison with standard keyframes are presented in \Sec\ref{sec:exp}.
We then conclude the paper with some discussion in \Sec\ref{sec:conclusion}.

%% file: sections/voxelhashing.tex
\begin{figure}
 \centering
 \includegraphics[scale=0.45]{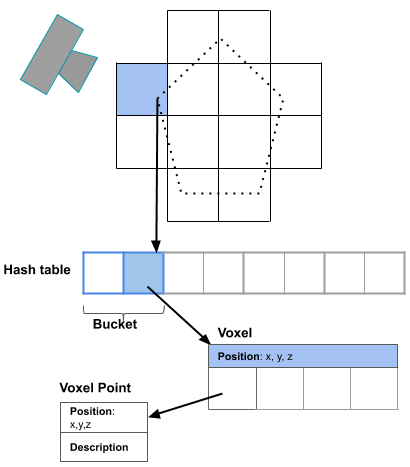} %
 \caption{Voxel hashing data structure.
  The map is stored as a hash table. The hash function maps the integer world coordinates to hash buckets.
  }
 \label{fig:voxel_hashing}
\end{figure}

\section{Voxel Hashing for SLAM} \label{data_struct}
In this section, we explain the data structure used to represent the voxel-map and how it can be integrated within a SLAM pipeline.

\subsection{Voxel hashing data structure}
The proposed voxel-map representation is based on the work of Nie{\ss}ner \etal \cite{Niessner13tog}.
It stores the map as a hash table as shown in \Fig \ref{fig:voxel_hashing}.
In particular, the world is comprised of voxels. Each allocated voxel stores its position in the world coordinates, and a list of voxel points (\ie actual 3D points of interest, such as 3D landmarks in SLAM) that are within this voxel.
As for the voxel points, each of them has a position in the world coordinate and also a description.
This description is used for 2D-3D data association in the tracking process (\ie frame-to-map alignment).
For feature-based methods, the description of a point is its feature descriptor, whereas for direct methods, the description is the image patch around the location where the feature is extracted.

We use an efficient C++ implementation of hash table to manage the allocation and retrieval of voxels. 
Each entry in the hash table contains a pointer to an allocated voxel. 
The allocated voxels are accessed from the hash table using a hashing function of the voxels' world coordinates. 
Specifically, the hash value $H(\cdot)$ is computed as \cite{Niessner13tog, Teschner03vmv}:
\begin{equation}
    H(x, y, z) = ((\round{x} \cdot p_1) \otimes (\round{y} \cdot p_2) \otimes (\round{z} \cdot p_3))\;mod\;n,
\label{eq:hash_val}
\end{equation}
where $p_1$, $p_2$ and $p_3$ are large prime numbers, $\round{}$ the rounding operation, $\otimes$ the bit-wise XOR operator,  $mod$ the modulo operator, $n$ the hash table size

The goal of voxel hashing is to manage 3D points efficiently. Combining the voxel representation and the hashing method, we can get the points inside a given region (\eg the camera viewing frustum) in constant time, regardless of the map size.
Alternatives, such as raw point clouds and keyframes, do not scale well as the map size increases.
For unstructured point clouds, each point needs to be checked individually to determine whether it falls in the area of interest.
Using overlapping keyframes is often a good heuristic, but one still needs to exhaustively check the points in these keyframes, since there is no guarantee that an arbitrary point in an overlapping keyframe will fall in the viewing frustum of the current frame.
In other words, keyframes and voxels can be both viewed as proxies of the underlying 3D points of interest, but using hash values based on the voxel positions enable us to directly get points in a precisely defined 3D area of interest, whereas overlapping keyframes can only rely on the weak covisibility assumption.
With the voxel-map, several technical details need to be taken care of:
\newline
\textbf{Voxel size}:
An appropriate voxel size has to be chosen for efficient performance.
In the extreme cases, if one point occupies a tiny voxel or all the points live in one voxel, the voxel map representation is no different from raw point clouds.
\newline \textbf{Resolving collisions}:
Due to the nature of the hash function, collisions can occur if multiple voxels get mapped to the same hash value.
To handle these collisions, we divide the hash table into buckets.
Each bucket corresponds to a unique hash value. These buckets store the hash entries as a list.
In the event of a collision, we accommodate the new voxel by adding the pointer to the corresponding bucket list.
With a reasonable selection of hash table size and bucket size, the collisions can be kept minimum.

\subsection{SLAM map management with voxels}
In general, a map in SLAM stores 3D geometric objects, such as points and lines, against which a new frame can localize.
A map is updated over time (\eg add/delete points, update the information of existing points), and should support efficient query in the tracking process (\eg what are the possibly matched points in a newly coming image?).
Below, we describe the corresponding functionalities in our voxel-map.
Note that we do not discard keyframes completely, as keyframe-based BA is still necessary for optimizing the map. Our voxel-map is instead a more efficient organization of the 3D points to facilitate data association.

\noindent
\textbf{Insert point}: We find the target bucket in the hash table using the hash function \eqref{eq:hash_val} on the world coordinates of the point.
We then iterate over the hash entries in the bucket. If a voxel exists in the space of the point, we add the point to the voxel.
If the position of the point is already occupied (\ie equality up to a certain precision), we update the description of the point with the newly added one.
If such a voxel does not exist, a new voxel is created and added as a hash entry to the bucket; the point is then added to the newly created voxel.

\noindent
\textbf{Delete point}: Deletion is performed similar to insertion.
We first calculate the hash value for the point position and then iterate over the hash entries in the target bucket till we find the voxel that fits the point position.
The point is then deleted from the voxel.

\noindent
\textbf{Query map}: 
To query 3D points at a given location, we first convert the world coordinates of the query location to integer world coordinates then compute the hash value.
If there exists a bucket corresponding to this hash value, we iterate over all the hash entries in the bucket till we find the voxel that fits the query location.
If the hash entry exists, we return the pointer to the allocated voxel, which contains exactly the 3D points around the query location. We will describe next how this strategy can be used to query possible visible points from a given pose.

\subsection{Point query with raycasting}\label{raycast}
\begin{figure}
 \centering
 \begin{subfigure}{0.4\linewidth}
 \includegraphics[width=\linewidth]{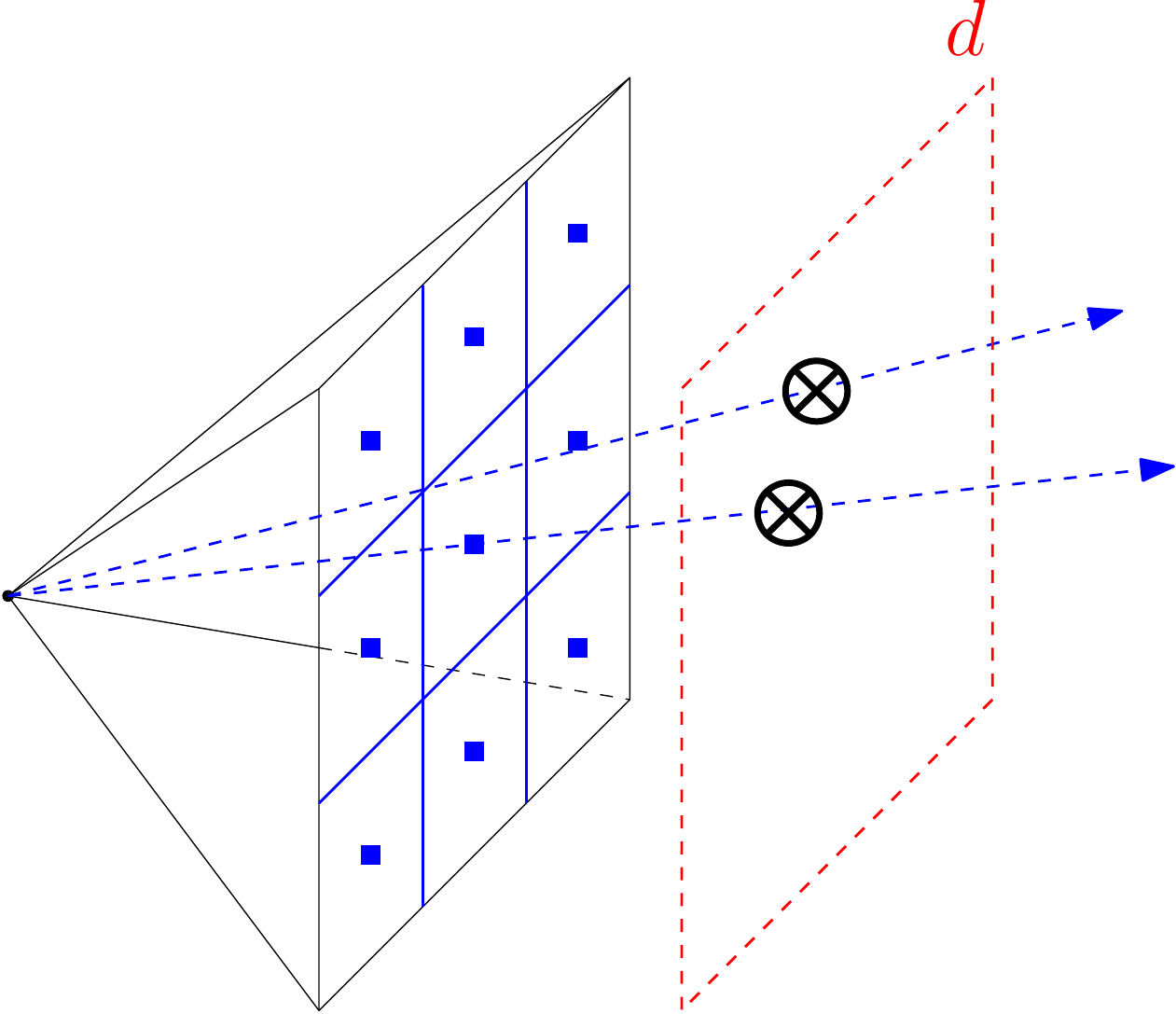}
 \end{subfigure}
 \begin{subfigure}{0.58\linewidth}
 \includegraphics[width=\linewidth]{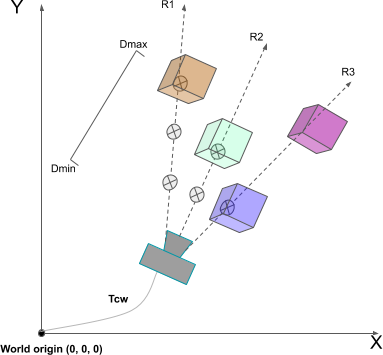}
 \end{subfigure}
 \caption{
 \textbf{Left}: Sampled pixels (blue) on the image plane are backprojected to rays in 3D, which are then sampled at a discrete set of distances (only two rays and one distance plane, $d$, is shown here to avoid cluttering).
 \textbf{Right}: Points query with raycasting.
 Here the voxels are colored only for the ease of illustration.
 }
 \label{fig:raycast_pts}
\end{figure}

A necessary function in the tracking process in SLAM is to query the points that is possibly visible in the current frame.
This is essentially the same as getting all the points that fall in the camera FoV (up to a cut-off distance).
Since our hashing function is based on the voxel position, we directly sample the frustum in a raycasting manner.
In particular:
\begin{itemize}
    \item 
    \textbf{Image plane sampling and raycasting}: 
    We first sample pixels from a regular grid on the image plane.
    Then we backproject these sampled pixels to bearing vectors in 3D space, resulting in $r$ rays $\{R_i\}_{i=1}^{r}$.
    These rays essentially samples the camera FoV.
    Note that the rays are expressed in the camera frame.
    \item
    \textbf{Ray sampling}: 
    For each ray $R_i$, We sample $s$ points $\{{}^{c}S_{j}^{i}\}_{j=1}^{s}$ from a depth range of $D_{min}$ to $D_{max}$.
    These points are again expressed in the camera frame $c$, which can be precomputed offline.
    \item
    \textbf{Points query}:
    In the context of SLAM, at the query time, we usually have a prior of the current camera pose (\eg previous camera pose or IMU propagation).
    With this prior, we can transform the aforementioned sample points to the world frame ${}^{w}S_{j}^{i}$ ($i=[1, r],\;j=[1, s]$),
    and query the voxels at these sample points using the hashing method in \Sec\ref{data_struct}.
\end{itemize}
The first two steps (shown in \Fig\ref{fig:raycast_pts} left) essentially discretize the camera frustum in the given distance range. Since the discretization is done in the camera frame, it only needs to be computed \textit{once} (possibly offline).
The time complexity of our approach is $\mathcal{O}(r\cdot s)$.
Since the $r$ and $s$ are independent of the map size, the query time remains constant, even as the map grows. In contrast, the map query time for keyframe-based map is $\mathcal{O}(k)$, where $k$ is the number of keyframes and increases with the map size.

In addition, we would like to emphasize that the sampling order for each ray is from the closest voxel to the most distant.
In this way, when querying the map along each ray, we only return the first occurring voxel, thus avoiding the occluded voxels along the same ray, as illustrated in \Fig\ref{fig:raycast_pts}.

%% file: sections/systemoverview.tex
\section{Case Study: SVO with voxel map}\label{algo}

To demonstrate the practical value of our map representation, we adapt a state-of-the-art keyframe-based sparse SLAM pipeline SVO~\cite{Forster17troSVO} to use the voxel-hashing based map and raycasting points query.
It is worth-noting that the proposed method is a general building block for SLAM, and widely applicable.

SVO is a hybrid pipeline that combines the advantages of direct and feature-based methods.
In particular, it first aligns the new image with the previous image by minimizing the photometric error over a sparse set of patches with known depths.
This gives a good prior about the pose of the new frame.
With the given prior, the pipeline finds the keyframes that overlap with the new frame.
The overlapping keyframes are found by projecting selected points from the current frame to the keyframes in the local map (ordered by distance to current keyframe) until a set of \textit{M} overlapping keyframes  are found. Since in most translation motion cases the closest \textit{M} keyframes with overlap are the latest \textit{M} keyframes, the average query time depends only on the value of \textit{M} and not on the map size.
The pipeline further searches for matches in the new image for the points from these keyframes by Lucas-Kanade tracking \cite{Baker04ijcv}.
Once the correspondences are established, the pose is estimated through a motion-only bundle adjustment.
The pipeline also has a separate mapping thread, which uses a robust Bayesian filter \cite{Vogiatzis11jivc} for depth estimation.

To integrate our proposed voxel-hashing map with SVO, we make the following adaptations:

\noindent
\textbf{Map query for motion estimation}:
The pose estimate from sparse image alignment gives a prior to find potential correspondences.
Instead of checking overlapping keyframes, we directly sample the camera frustum as described in Section \ref{raycast}.
Our raycasting-based map query return a set of voxels that are visible, without occluded voxels, from the camera frustum.
Here we assume points from the same voxel do not occlude each other, which is mostly true considering the sparsity of the map.
Then the points within these voxels are tracked as in the original pipeline.

\noindent
\textbf{Map management}:
If the depth uncertainty for a point reduce to a certain threshold in the depth filter, a new 3D point is initialized at the estimated depth.
This point is then inserted in the voxel map as described in Section \ref{data_struct} and used for camera pose tracking afterwards. 
Moreover, during the tracking stage, certain points are labeled as outliers (\eg in the motion-only bundle adjustment).
These points are removed from the voxel map as described in Section \ref{data_struct}.

%% file: sections/experiments.tex
\section{EXPERIMENTAL EVALUATION}
\label{sec:exp}

We organize our experimental results in two parts: simulation and real-world experiments.
In simulation, we compared the performance of our voxel-map and raycasting map query with a naive-keyframe-based method, in terms of map query time and occlusion handling.
We also performed experiments on EuRoC \cite{Burri15ijrr} dataset, where we focused on the pose estimation accuracy of overlapping keyframe map, as described in \Sec\ref{algo} and our proposed approach.

\begin{figure}
 \centering
 \includegraphics[width=0.48\textwidth]{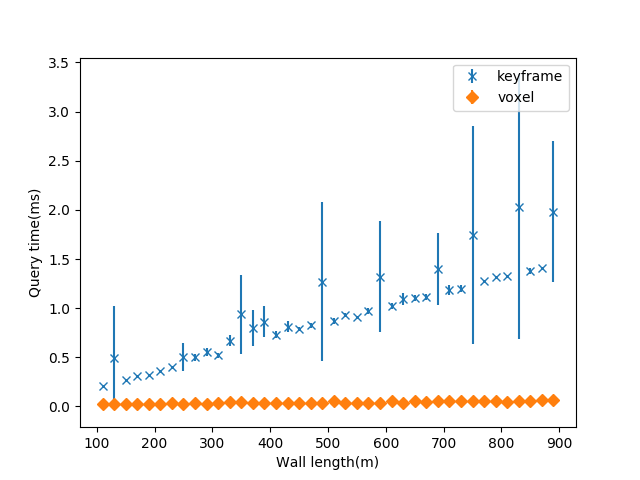}
 \caption{Comparison of time taken for querying the map as the map size increases.}
 \label{fig:map_query_time_simulation}
\end{figure}

\subsection{Simulation}
\label{subsec:sim}
\textbf{Map query time}: The goal of this experiment is to show that our method scales better than keyframes as the map size grows.
To this purpose, we simulated a map that consists of a straight wall.
We then queried the map (\ie get visible 3D points) at 10 locations along a line parallel to the wall.
For different map size, we increased the length of the wall from 100\si{m} to 900\si{m}. To ensure same density of map points, map points increased from 1000 to 9000.
We established two different map representations:
\newline
\textbf{Naive-Keyframe}:
We sampled keyframes evenly on the wall, so that each point belongs to a unique keyframe.
The maximum number of points in each keyframe was fixed to $100$.
Therefore, as the length of the wall increases, the number of keyframes in the map also increases.
This is to mimic an exploration scenario, where the map keeps expanding.
To query the visible points at a given pose, we iterated over the points in the keyframes, and once there is one point visible from the query pose, we considered this keyframe overlapping with the query pose and continued to the next keyframe.
The query time was the total time of checking all the keyframes.
\newline
\textbf{Voxel-hashing}:
We allocated sufficient voxels to hold all the map points.
The voxel grid size was fixed to 2\si{m}.
At query time, we used the raycasting based method described in \Sec\ref{raycast} to return a list of visible points.

\begin{figure}[ht!]
	\centering
	\begin{subfigure}{0.8\linewidth}
		\includegraphics[scale=.4]{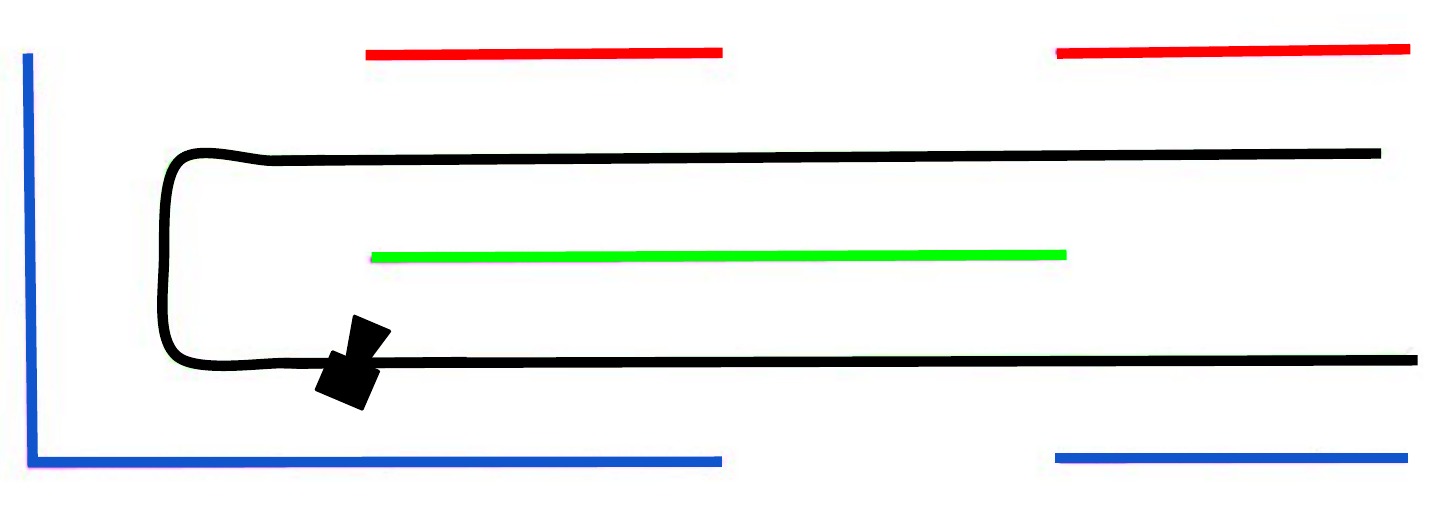}
		\caption{}
		\label{map_corridor}
	\end{subfigure}
	
	\begin{subfigure}{.4\linewidth}
		\includegraphics[width=\linewidth]{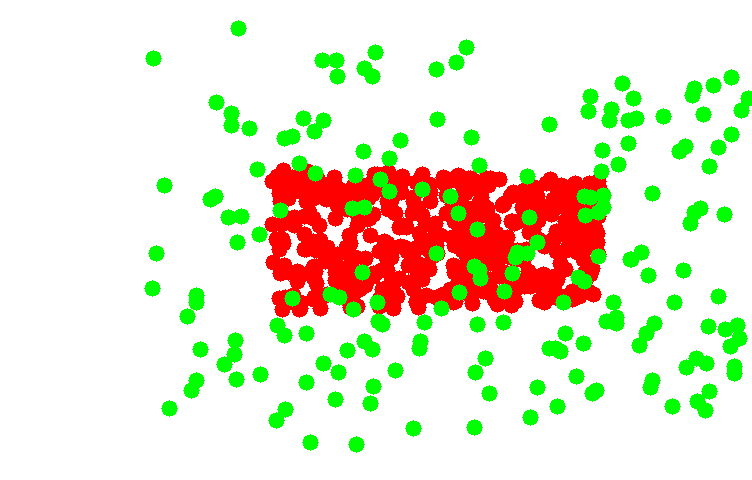}
		\caption{}
		\label{keyframe_sim}
	\end{subfigure}
	\hskip2em
	\begin{subfigure}{.4\linewidth}
		\includegraphics[width=\linewidth]{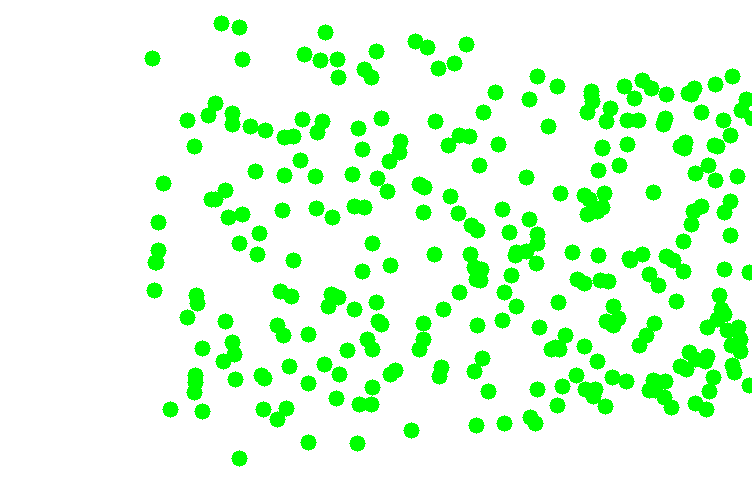}
		\caption{}
		\label{voxel_sim}
	\end{subfigure}
	\caption{The top view of the simulation environment to test map query in the presence of an occlusion \ref{map_corridor}. The map consists of a long U-shaped corridor. The walls in the map are colored by distance from the camera with red being far away , green closer and blue outside the FoV. The camera is moving along a trajectory (black) while changing viewing direction. \Fig\ref{keyframe_sim} shows map query using keyframe map. \Fig\ref{voxel_sim} shows map query using voxel map}
	\label{fig:map_sim}
\end{figure}

We compared the map query time for the above representations as the map sized varied.
Each experiment was repeated $5$ times.
The results, plotted in \Fig \ref{fig:map_query_time_simulation}, show that for a smaller map size, all methods have similar query time.
However, as the map size increased, the query time for the keyframe-based method increased almost linearly;
in contrast, the query time using the proposed method remains constant.
This makes our method particularly useful to manage map points at a large scale.

\textbf{Geometric awareness}: In this experiment, we were interested in validating the occlusion handling capability of our method.
We simulated a corridor-like environment and a camera trajectory through the corridor, as shown in \Fig\ref{map_corridor}.
The naive-keyframe-based and voxel-map were generated in a similar manner as in the previous simulation.
We then queried the map for visible points from the poses along the camera trajectory at the point of turn(as shown in \Fig\ref{map_corridor}).
The visible points were then projected into the image plane for visualization, the color of the points indicates distance from camera (red being farther away and green being closer).
The camera was looking at two planes of points at different distances.
Ideally, the points on the plane at the farther distance(\ie red points) should be occluded by the nearer ones (\ie green points).
While the naive-keyframe query had no notion of occlusion (\Fig\ref{keyframe_sim}); our method, thanks to the raycasting query scheme, was able to recognize the farther points along the same ray as occluded (\Fig\ref{voxel_sim}).

\begin{table*}[t]
\tiny
\caption{RMSE(m) original error on tested sequences from the EuRoC datasets. 
In bold is the best value in the column and the underlined is the second best value. 
For the voxel map, the blue color indicates it performed better than KF$5$, red indicates the worst performance.}
\label{rmse}
\begin{center}
\resizebox{\textwidth}{!}{
\begin{tabular}{c|c c c c c c c c c }
\hline
     \textbf{Algorithm} &  \textbf{MH\_01} & \textbf{MH\_02} & \textbf{MH\_03} & \textbf{MH\_04} & \textbf{ MH\_05} & \textbf{V1\_01} & \textbf{V1\_02 }& \textbf{V2\_01} &  \textbf{V2\_02} \\
\hline
\textbf{Ours}& \improv{0.120}  & \textbf{\improv{0.083}}  & \improv{0.856} & \red{2.575} & \underline{\improv{0.902}} & \textbf{\improv{0.266}} & \textbf{\improv{0.686}} & \improv{0.336} & \improv{0.825} \\
\textbf{KF5}   & 0.255 & 0.432 & 1.968 & \textbf{1.315} & 0.930 & 0.706 & 1.130 & 0.709 & 1.060\\
\textbf{KF10}  & \textbf{0.085} & 0.474 & 0.765 & \underline{1.343} & 1.081 & 0.516 & 0.974 & 0.356 & \textbf{0.711}\\
\textbf{KF15}  & 0.094 & 0.167 & 0.852 & 1.393 & 0.989 & 0.647 & \underline{0.876} & 0.281 & 0.849\\
\textbf{KF20}  & \underline{0.092} & \underline{0.155} & \underline{0.712} & 1.552 & 0.908 & 0.553 & 0.959 & \textbf{0.253} & 0.842\\
\textbf{KF25}  & 0.117 & 0.251 & 0.726 & 1.829 & 0.895 & 0.558 & 0.996 & 0.307 & 0.867\\
\textbf{KF30}  & 0.144 & 0.367 & \textbf{0.671} & 2.042 &\textbf{0.791} & \underline{0.478} & 0.949 & \underline{0.259} & \underline{0.824}\\
\hline
\end{tabular}}
\end{center}
\end{table*}

\begin{table*}[t]
\tiny
\caption{Average total time (ms) for each frame. In bold is the best value in the column and underlined is the second best value.}
\label{time_euroc}
\begin{center}
\resizebox{\textwidth}{!}{
\begin{tabular}{c|c c c c c c c c c }
\hline
     \textbf{Algorithm} &  \textbf{MH\_01} & \textbf{MH\_02} & \textbf{MH\_03} & \textbf{MH\_04} & \textbf{ MH\_05} & \textbf{V1\_01} & \textbf{V1\_02 }  & \textbf{V2\_01} &  \textbf{V2\_02} \\
\hline
\textbf{Ours} & \underline{4.818} & \textbf{4.539} & \textbf{5.309} & \underline{5.208} & \underline{5.042} & \underline{4.096} & \underline{5.655} & \underline{3.850} & \underline{5.358}  \\
\textbf{KF5}  & \textbf{4.642} & \underline{4.690} & \underline{5.459} & \textbf{5.071} & \textbf{4.753} & \textbf{3.686} & \textbf{5.244} & \textbf{3.349} & \textbf{4.603}  \\
\textbf{KF10} & 5.732 & 5.882 & 6.429 & 5.708 & 5.980 & 4.418 & 6.044 & 3.968 & 5.401  \\ 
\textbf{KF15} & 6.266 & 6.139 & 7.004 & 6.545 & 6.289 & 4.722 & 6.167 & 4.234 & 5.780  \\
\textbf{KF20} & 6.950 & 6.766 & 7.995 & 7.035 & 6.290 & 4.968 & 6.560 & 4.551 & 6.065  \\
\textbf{KF25} & 6.976 & 6.957 & 8.049 & 6.390 & 6.564 & 5.232 & 6.811 & 4.784 & 6.502  \\
\textbf{KF30} & 7.229 & 7.477 & 8.534 & 6.560 & 6.656 & 5.435 & 7.202 & 5.054 & 6.832  \\
\hline
\end{tabular}
}
\end{center}
\end{table*}

\subsection{Real-world experiments}
\label{subsec: real_world_exp}
\begin{figure}[t!]
    \centering
    \begin{subfigure}{0.49\linewidth}
        \includegraphics[width=\linewidth]{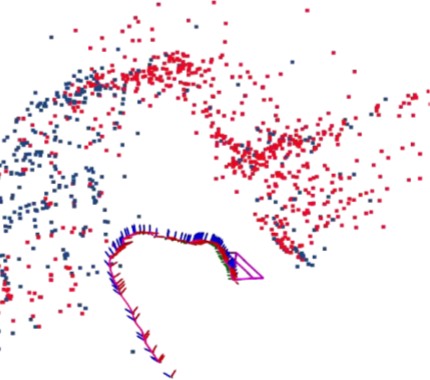}
        \caption{Map query using keyframes.}
        \label{kf_corner}
    \end{subfigure}
    \begin{subfigure}{0.49\linewidth}
        \includegraphics[width=\linewidth]{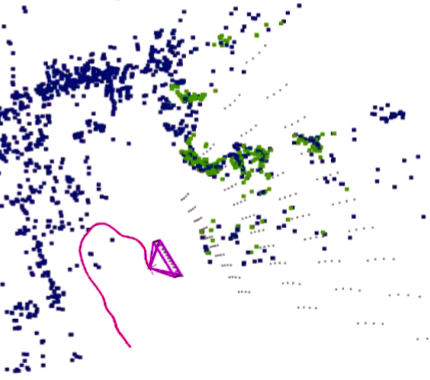}
        \caption{Map query using voxel map.}
        \label{voxel_corner}
    \end{subfigure}
    \caption{Results of querying keyframe-map and voxel-map at the same pose (purple pyramid) in a SLAM pipeline.
    Blue (both): all points in the map;
    Red: visible points queried from overlapping keyframes;
    Green: visible points queried from the proposed voxel-map;
    Gray: sample points in the raycasting-based query (\Sec\ref{raycast}).
    }
    \label{fig:mh01_vis}
\end{figure}

We ran monocular version of SVO in two configurations on EuRoC: 1) the original pipeline with an increase in the number of keyframes in map from $5$ to $30$ ; 2) the adapted pipeline described in \Sec\ref{algo}, which stores all the landmarks in the map, since this has no overhead in saving candidates in the map.
Note that we omitted the results on \textit{V1\textunderscore03} and \textit{V2\textunderscore03}, as the aggressiveness in these two sequences made both pipelines (vision-only) unstable so that no meaningful comparison could be made.
The Root Mean Squared Error (RMSE) on the tested sequences were computed using the evaluation protocol (\text{SIM}$3$ alignment with all the frames) in \cite{Zhang18iros}. The results are shown in Table.~\ref{rmse}.
We also computed the time required for the entire pipeline per frame of the dataset. The results are shown in Table~\ref{time_euroc}.
It can be seen  from Table ~\ref{rmse} that, as the map size increases from 5 to 30 keyframes, the RMSE reduces overall. 
The results here might vary from the reported values from \cite{Forster17troSVO} due to different parameters for the experiment.
Table ~\ref{time_euroc} shows that with an increase in the map size, the computation time increases. This shows the efficiency-accuracy trade-off in keyframe map in terms of computation time and estimation error. In contrast, the voxel map is able to get better accuracy for lower computation time. 
It does not always outperform all the keyframe sequences because the performance of the voxel map depends on the environment and the voxel size. In the case of \emph{MH\textunderscore04}, the voxel map  did not have a consistent scale across the trajectory, which caused higher RMSE.
We emphasize that our method performs similar to the lowest keyframe map size (5 keyframe map \ie KF$5$) in terms of computation time, while achieving a higher accuracy.
For example, we achieve an improvement ranging from 3\% (on \emph{MH\textunderscore05}) to 80\% (on \emph{MH\textunderscore02}), with an average of 46.2\%, as compared to KF$5$, on the EuRoC sequences. The color of the voxel map indicates the performance with respect to KF$5$, blue indicating the improvement of the voxel map over KF$5$ and red indicating the worst performance.
We also show the qualitative results in \emph{MH\textunderscore01} in \Fig\ref{fig:mh01_vis}, where we can see that the points returned by voxel map were much more consistent with the FoV of the camera.

We evaluated the effect of voxel size on the performance and computation time. We ran the same pipeline of voxel-based map with SVO on the dataset \emph{MH\textunderscore01}. We increased the voxel size gradually from 0.5\si{m} (extreme scenario where a voxel contains a very small number of points.) to 20\si{m} (extreme scenario where a voxels contain all the map points.) and calculated the inlier ratio, RMSE and average total time required by the pipeline per frame. 
The inlier ratio is computed as the ratio of successfully reprojected points and the total map points returned by the query. Therefore, higher inlier ratio indicates better map points were retrieved by the query.
The results in Table \ref{voxel_prms} show that the voxel size affects the estimation error as can be evident from the inlier ratio and RMSE.
For smaller voxel sizes, the voxel map acts as point cloud representation; however, due to discretization of the space, we miss many map points resulting in a lower inlier ratio and a higher RMSE. The time is also significantly higher as we have to search for many voxels along the rays from the frustrum, which increases the query time.
For extremely large voxel sizes, the inlier ratio is also low as we include some points that do not fall in the FoV of camera; while this affects the inlier ratio, the effect on RMSE is acceptable. 
With this experiment, we show that to get better estimation performance, we need to  choose an appropriate voxel size.

\begin{table}[t]
\small
\caption{Effect of the voxel size on RMSE of MH\_01 sequence. Bold values represent the lowest values in the columns and underlined are the second best values.}
\label{voxel_prms}
\begin{center}
\begin{tabular}{c|c c c}
\toprule
      \textbf{Voxel size(m)} & \textbf{Time(ms)} & \textbf{Inlier ratio} & \textbf{RMSE(m)}\\
\midrule
\textbf{0.5} & 9.286 & 0.378 & 0.778 \\
\textbf{5}   & 5.618 & \textbf{0.822} & 0.278 \\
\textbf{10}  & 5.767 & \underline{0.762} & \textbf{0.217}\\
\textbf{15}  & \textbf{5.378} & 0.721 & \underline{0.225}\\
\textbf{20}  & \underline{5.551} & 0.715 & 0.499\\
\bottomrule
\end{tabular}
\end{center}
\end{table}

\subsection{Discussion}
Using the voxel-map brings better geometric reasoning than keyframes, as shown in \Sec\ref{subsec:sim}.
Moreover, it is possible to further generate more complicated/useful geometric representations from a regular voxel grid (\eg distance field for planning \cite{Oleynikova17iros}), which opens the door to better geometric reasoning for sparse SLAM.
Although the focus of this work is not about accuracy, the scalability of our method could benefit the accuracy of SLAM pipelines as well.
In general, being able to maintain and query a larger map brings better accuracy for SLAM pipelines.
In practice, however, this is usually limited by the computational power for real-time applications.
Therefore, we believe our method, due to the constant query time (regardless of map size), is a valuable tool for accurate SLAM pipelines, as shown in \Sec\ref{subsec: real_world_exp}.
We believe that this approach is a step towards the direction of using maximally efficient map representations (in terms of accuracy, efficiency and geometry awareness) for SLAM.

%% file: sections/conclusion.tex
\section{Conclusions and Future Work}
\label{sec:conclusion}
In this paper, we proposed a scalable and geometry-aware voxel-map for sparse SLAM, aiming to replace keyframes for data association in the tracking process.
The map is organized in voxels, and each voxel can be accessed in constant time using a hashing function on its location.
Using the voxel-hashing method, visible points from a camera pose can be efficiently queried by sampling the camera frustum in constant time, which makes the proposed method scale well with large scenes.
Moreover, by sampling the frustum in a raycasting fashion, we were able to handle occlusion, which is not possible using keyframes.
We validated the advantages of the proposed method over keyframes using simulation as well as on real-world data with a modern visual SLAM pipeline.

As future work, we would like to explore the use of the voxel-map in non-trivial multi-camera configurations, where the map management with keyframes becomes complicated.
How to efficiently generate high-level geometric structure, such as meshes and dense surface, from the voxel map is also of interest for robotic applications such as motion planning.